\documentclass[runningheads]{llncs}
\usepackage[T1]{fontenc}
\usepackage{graphicx}
\usepackage{hyperref}
\usepackage{verbatim}
\usepackage{siunitx}
\usepackage{subcaption}
\begin{document}
\title{Fiber-Resolved Microstructure Quantification from Multi-Shell Diffusion MRI using Detection Transformers}
\titlerunning{Fiber-Resolved Microstructure Quantification using Detection Transformers}
\newcommand{\equalcontrib}{\textsuperscript{*}}
\newcommand{\github}{\href{https://github.com/Marcus02W/Diffusion-DETR}{github.com/Marcus02W/Diffusion-DETR}}
\author{Sebastian Endt\equalcontrib\inst{1,2}\orcidID{0000-0003-2062-936X} \and
Marcus Wirth\equalcontrib\inst{1} \orcidID{0009-0000-6567-395X} \and
Johannes R. Schlund\inst{1}\orcidID{0009-0009-1792-6807} \and
Marion I. Menzel\inst{1,3}\orcidID{0000-0003-0087-9134}}
\authorrunning{S. Endt et al.}
%
\institute{AImotion Bavaria, Technische Hochschule Ingolstadt, Ingolstadt, Germany \email{sebastian.endt@thi.de} \and
TUM School of Computation, Information and Technology, Technical University of Munich, Garching, Germany \and
TUM School of Natural Sciences, Technical University of Munich, Garching, Germany}
\maketitle
\begingroup
    \renewcommand{\thefootnote}{}
    \footnotetext{\equalcontrib{} Authors contributed equally to this work.}
\endgroup
\begin{abstract}
Fiber orientation and compartmental microstructure are central to the characterization of white matter tissue in diffusion MRI, yet existing methods either resolve fiber orientations without quantifying microstructure, or quantify microstructure while assuming a fixed number of compartments and a single fiber direction. Nonparametric approaches that recover both require tensor-valued diffusion encoding and computationally expensive Monte-Carlo inversion of an ill-posed inverse Laplace transform. We propose to reframe this problem as an object detection-like task, adopting the Detection Transformer (DETR) architecture to jointly predict mean diffusivity (MD), fractional anisotropy (FA), main fiber direction, and signal fraction for a variable number of compartments per voxel from standard multi-shell diffusion MRI with linear encoding. Hungarian matching during training resolves permutation invariance across compartments. We introduce mean Average Precision as a reproducible benchmark metric. Evaluated on synthetic test data with up to five compartments per voxel, our model achieves $R^2=0.95$ for MD, $R^2=0.88$ for FA, and a median angular error of $\SI{4.2}{\degree}$, with performance scaling naturally with compartmental signal fraction.

All code is publicly available at \github.


\end{abstract}
\section{Introduction}
Conventional MRI resolves a multitude of different contrasts, but imaging always comes with limited resolution. More specific methods, i.e. diffusion imaging\index{diffusion imaging}, indirectly provide information about microstructure. However they do not quantify the multiple microstructure compartments.

Diffusion tensor imaging\index{diffusion tensor imaging} (DTI) is limited to a single diffusion tensor per voxel and cannot resolve regions with complex fiber configurations, like crossing nerve fibers\index{crossing fibers}, which are very common in human white matter \cite{jeurissen2013investigating}. Parametric tissue models like NODDI \cite{zhang2012noddi} and DIAMOND \cite{scherrer2016characterizing} assume a fixed number of compartments, usually with one single fiber direction, and may fail to generalize to unexpected tissue structures. Methods to recover the fiber distribution (bedpostX, CSD, MSMT-CSD) do not resolve other compartment specific parameters \cite{behrens2007probabilistic,tournier2007robust,JEURISSEN2014411}.

In conclusion, compartmental microstructure and fiber orientations are typically estimated sequentially and independently. No method with standard diffusion encoding jointly recovers both data.

Multicomponent imaging\index{multicomponent imaging} and multiparametric correlation imaging\index{multiparametric correlation imaging} recover microstructure compartments without assuming a biological model by linking a rich multi-contrast data set to distributions (spectra) of tissue parameters. Many works focus on relaxometry and vary inversion time or echo time, to recover voxel wise T1/T2 spectra \cite{MacKay.1994,kim2020,Nagtegaal.2023}. The inclusion of diffusion weighting principally adds information about nerve fibers, but blows up the dimensionality of the spectra. The already ill-posed underlying inverse Laplace transform\index{inverse Laplace transform} (ILT) becomes even more challenging to solve. Usually correlation imaging studies with diffusion derive apparent diffusion coefficients, but provide no information about compartmental diffusion direction \cite{Avram.2019,Slator.2019,Avram.2021,Naranjo.2021,Luo.2023,Endt.2023,Hu.2024,Liu.2024,Xu.2024,Zong.2024,Ning.2025}.

Existing nonparametric works that resolve both fiber orientation and microstructure use computationally expensive inversions and tensor-valued diffusion encoding with different b-tensors \cite{REYMBAUT2021101988,dealmeidamartins2021computing,martin2021nonparametric,reymbaut2021toward,Rosenberg2022}. MC-SHORE quantifies microstructure including ODFs, but requires additional T1 encoding to separate pre-defined compartment types \cite{BOGUSZ2025110998}.

Learning-based methods have been applied to relaxation spectrum reconstruction \cite{yu2021,endt2021unmixing}, to microstructure estimation from diffusion MRI \cite{golkov2016,reisert2017}, and more recently to jointly recovering fiber orientations and compartmental parameters from standard multi-shell data \cite{dessain2024,consagra2025}. However, most existing approaches either assume a fixed number of compartments or require a separate fiber orientation step.

We propose the joint recovery of fiber orientation (main diffusion direction) and other compartmental diffusion metrics (mean diffusivity (MD), fractional anisotropy (FA), signal fraction) from standard multi-shell diffusion data without solving an ILT.

To that end, we interpret the 5D diffusion-diffusion correlation problem as an object detection\index{object detection} task. The unknown number and nature of compartments relate directly to a set prediction problem. We adopt the Detection Transformer (DETR)\index{detection transformer} by Carion et al. with learned queries to predict a variable number of compartments in a single forward pass, with Hungarian matching\index{hungarian matching} resolving permutation invariance \cite{carion2020}. This allows us to jointly estimate MD, FA, main direction, and signal fraction for each compartment.

\section{Methods}
\subsection{Data generation}
\begin{figure}
    \centering
    \includegraphics[width=\textwidth]{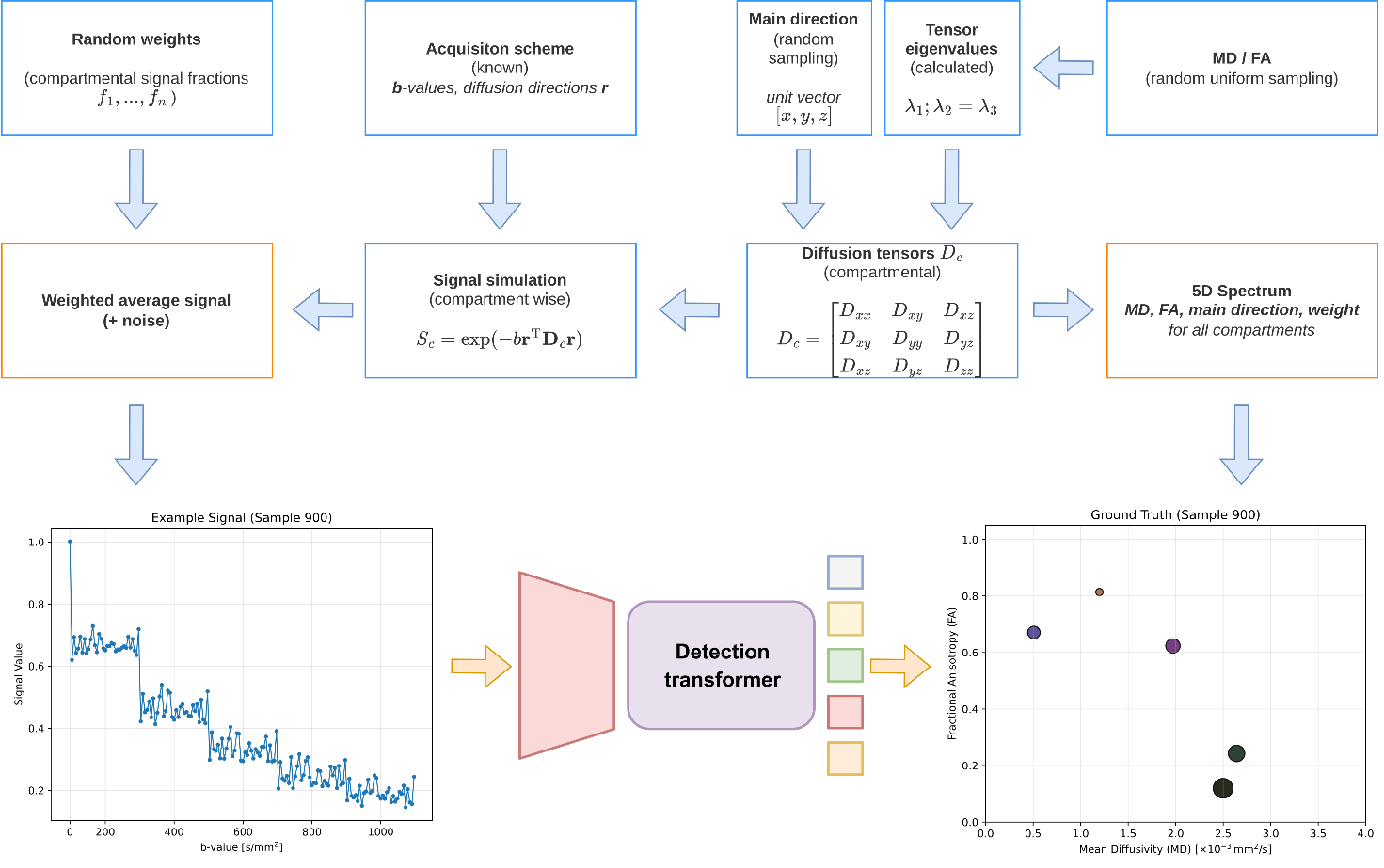}
    \caption{Pipeline for the generation of our synthetic data set \cite{schlund2025}. Starting from randomly generated MD, FA, and main direction, diffusion tensors for each compartment are calculated. Based on these, compartmental signal curves following a given acquisition scheme are simulated. The signals of all compartments in a sample are averaged according to random signal fractions before noise is added. The final signal curve is then paired with the set of MD, FA, and main direction for every compartment in the respective sample, to form a paired data set for supervised training.}
    \label{fig:pipeline}
\end{figure}

For supervised network training and evaluation against a known ground truth, we simulate a synthetic multi-compartment data set, consisting of sets of compartmental MD, FA, main direction, and weight, and the corresponding multi-shell signal curves (see fig. \ref{fig:pipeline}). The number of compartments per sample varies from $n_c=2$ to $n_c=5$.

Compartmental MD and FA are randomly sampled from uniform distributions: MD $\in[0, \SI{4e-3}{\square\milli\meter\per\second}]$, FA $\in[0, 1]$. Directions were sampled uniformly on the unit hemisphere, normalized to unit length, and projected to the $x\geq0$ hemisphere. Depending on the number of compartments $n_c$, weights were randomly generated, while ensuring a sum of $1$.

Under the assumption that the eigenvalues $\lambda_\perp:=\lambda_2=\lambda_3$ are equal, compartmental diffusion tensors were calculated from the generated MD, FA, and direction. From the tensors, compartmental diffusion signals were simulated using the acquisition scheme from \cite{hansen2016} up to $b=\SI{1000}{\second\per\square\milli\meter}$, resulting in 166 diffusion weightings \cite{hansendata}. The $n_c$ signal curves of a sample were weighted according to the respective signal fraction and averaged. Finally, $\SI{1}{\percent}$ random Gaussian noise relative to the $b=0$-signal was added.

In total $250'000$ samples are generated and split into training, validation and test data in a ratio of 80-10-10.

\subsection{DETR architecture}
\begin{figure}
    \centering
    \includegraphics[width=0.75\textwidth]{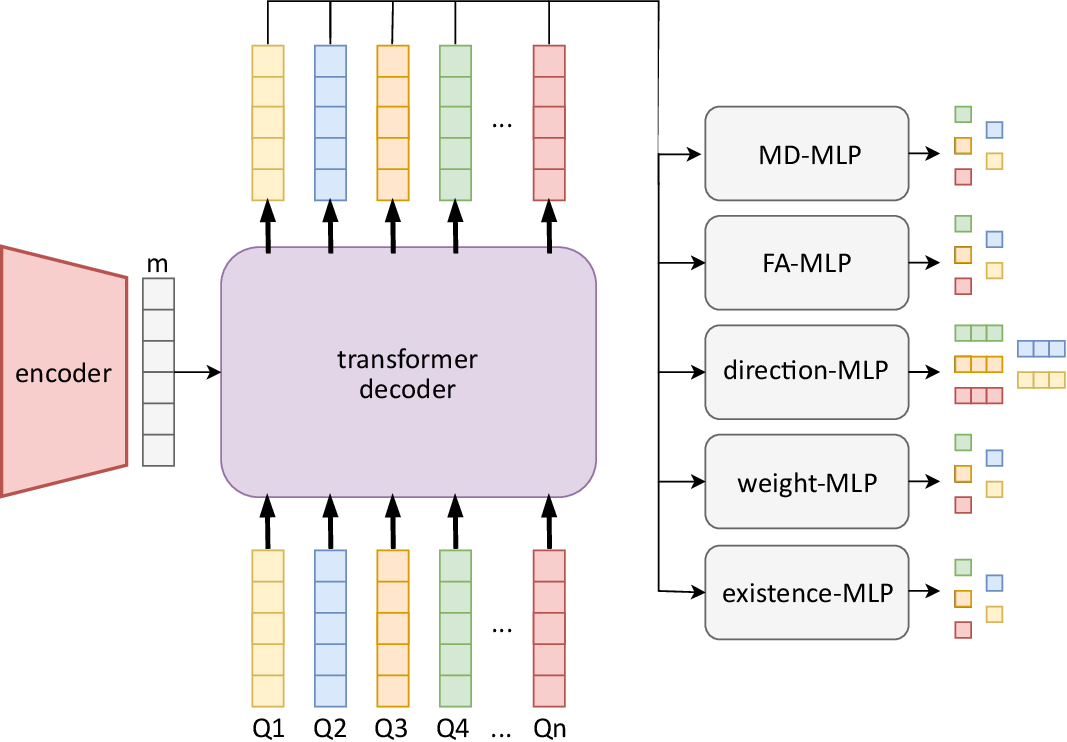}
    \caption{Similar to the original DETR model, the architecture consists of an encoder that produces a hidden vector, which is processed by a transformer decoder that uses cross-attention to fuse this information with the learned query vectors $Q$. The resulting representations per query are then fed to the individual prediction heads one by one. Only the existence head concatenates all queries into a single vector and outputs a vector containing the scores for all queries.}
    \label{fig:model}
\end{figure}

We adapt the DETR (DEtection TRansformer) architecture \cite{carion2020} to predict a variable number of discrete compartments in the 5D MD/FA/direction-spectrum. An MLP encoder processes the input signal into a hidden representation, which is combined with learned object queries in the transformer decoder via cross-attention (cf. Fig.~\ref{fig:model}). Hungarian matching guides training toward a globally optimal assignment between predicted and ground-truth compartments \cite{Hungarian}.

We use dedicated regression heads for MD, FA, direction, and weight. Since all prediction targets are continuous and thus only regression heads are present within the model, a No Object class as found in the original DETR model cannot be directly added to any prediction head. Thus, an additional existence head is introduced to determine compartment presence. MSE loss is used for MD and FA; MAPE loss for weight, ensuring equal treatment across weight scales. The direction loss accounts for antipodal orientation equivalence and is weighted by ground-truth FA and weight:
\begin{equation}
    \mathcal{L}_{\text{D}} = \frac{1}{M} \sum_{i=1}^{M} \left( 
    \frac{\min(\theta_i, 180^\circ - \theta_i)}{90^\circ} \right) \cdot 
    \hat{FA}_i \cdot \hat{W}_i, \label{eq:direction}
\end{equation}
where $\theta_i$ is the angular difference for compartment $i$ and $M$ the total number of compartments. Focal loss \cite{focal_loss} is used for the existence head, handling the class imbalance between the 40 queries and at most 5 true compartments per sample.

\subsection{Evaluation framework}
We adopt mean Average Precision\index{mean Average Precision} (mAP) \cite{pascalvoc}, a standard object detection metric, computed as the 101-point interpolation of the Precision-Recall curve \cite{mscoco}. To define true positives (TPs), we replace the standard bounding-box IoU with a dimension-wise relative error across MD, FA, and direction, normalized by their respective value ranges ($\SI{4e-3}{\square\milli\meter\per\second}$/s, $1.0$, and $\SI{180}{\degree}$). A prediction is counted as a TP if this relative error is below $10\%$ in all three dimensions simultaneously. We term this threshold-based metric mAP@10. If multiple predictions match the same ground-truth compartment, only the closest one is retained as TP and the rest are marked as FP, penalizing over-assignment.

For evaluations besides mAP, the existence score threshold for filtering predicted queries is chosen by taking the existence score resulting in the best validation data F1-Score as the final model-specific threshold.
\section{Results}
\begin{figure}[t]
    \centering
    \begin{subfigure}[t]{0.31\textwidth}
        \centering
        \includegraphics[width=\linewidth]{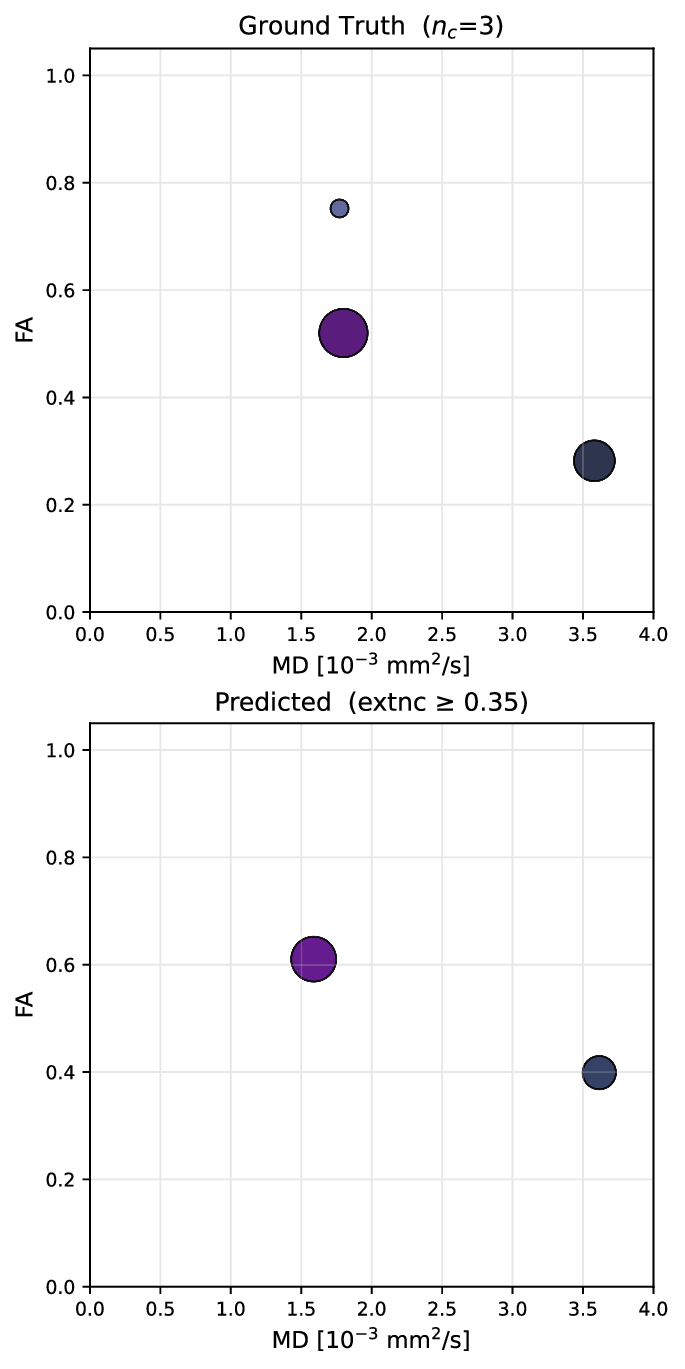}
        \caption{$n_c = 3$}
        \label{sfig:spectrum_nc3}
    \end{subfigure}
    \hfill
    \begin{subfigure}[t]{0.31\textwidth}
        \centering
        \includegraphics[width=\linewidth]{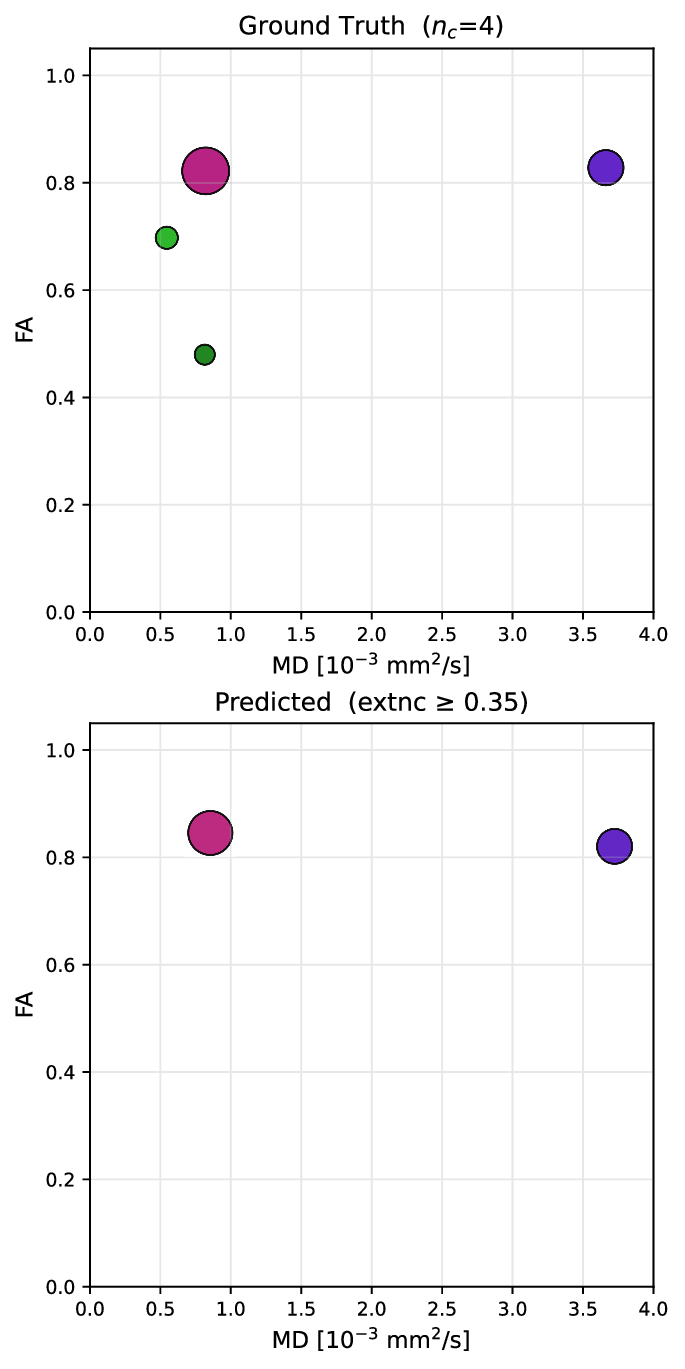}
        \caption{$n_c = 4$}
        \label{sfig:spectrum_nc4}
    \end{subfigure}
    \hfill
    \begin{subfigure}[t]{0.31\textwidth}
        \centering
        \includegraphics[width=\linewidth]{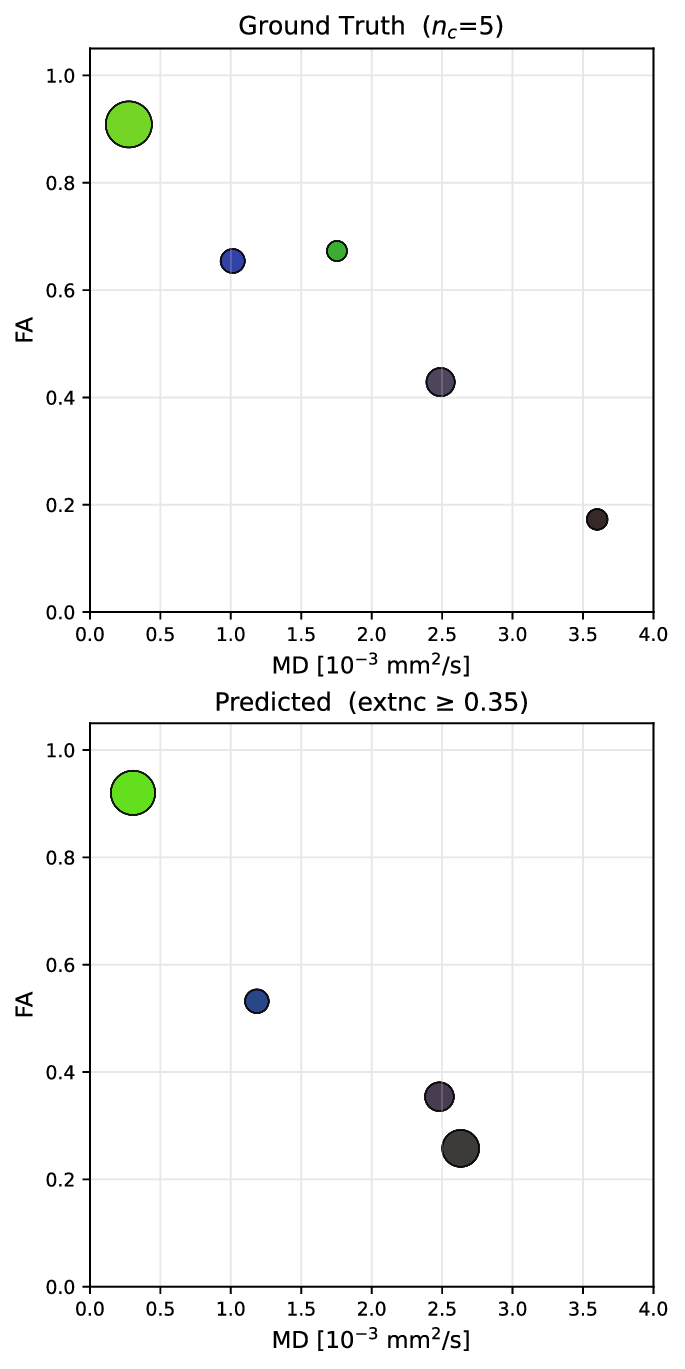}
        \caption{$n_c = 5$}
        \label{sfig:spectrum_nc5}
    \end{subfigure}
    \caption{Exemplary spectra for samples with 3, 4, and 5 compartments (left to right). Each subfigure shows the ground-truth spectrum (top) and our prediction (bottom) in the MD/FA plane. Marker size encodes signal fraction; color encodes the FA-weighted principal direction.}
    \label{fig:spectra}
\end{figure}

Fig. \ref{fig:spectra} shows random examples of predicted spectra and their corresponding ground truths at the determined optimal existence threshold of $0.35$. Predictions are generally better for compartments with higher FA and especially higher signal fractions.  Naturally, samples with $n_c=5$ compartments (Fig. \ref{sfig:spectrum_nc5}) prove to be more challenging than samples with low $n_c$.

The extent to which reconstruction performance depends on compartment weights is illustrated in Tab. \ref{tab:map}. Despite the considerably higher representation of small compartments, they are predicted less accurately. On the other hand, large compartments are correctly reconstructed with high mAP.

\begin{figure}[t]
    \centering
    \begin{subfigure}[t]{0.37\textwidth}
        \centering
        \includegraphics[width=\linewidth]{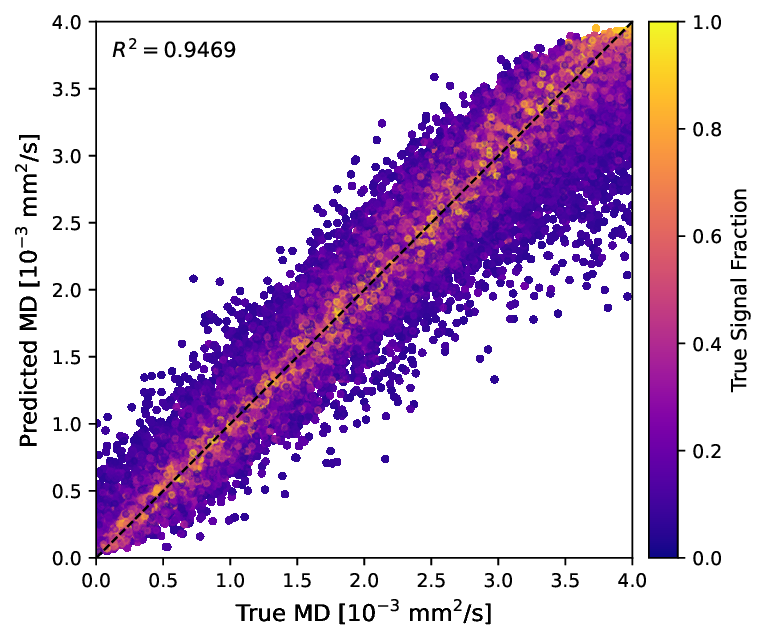}
        \caption{Predicted vs. true MD}
        \label{sfig:scatter_md}
    \end{subfigure}
    \hspace{0.015\textwidth}
    \begin{subfigure}[t]{0.37\textwidth}
        \centering
        \includegraphics[width=\linewidth]{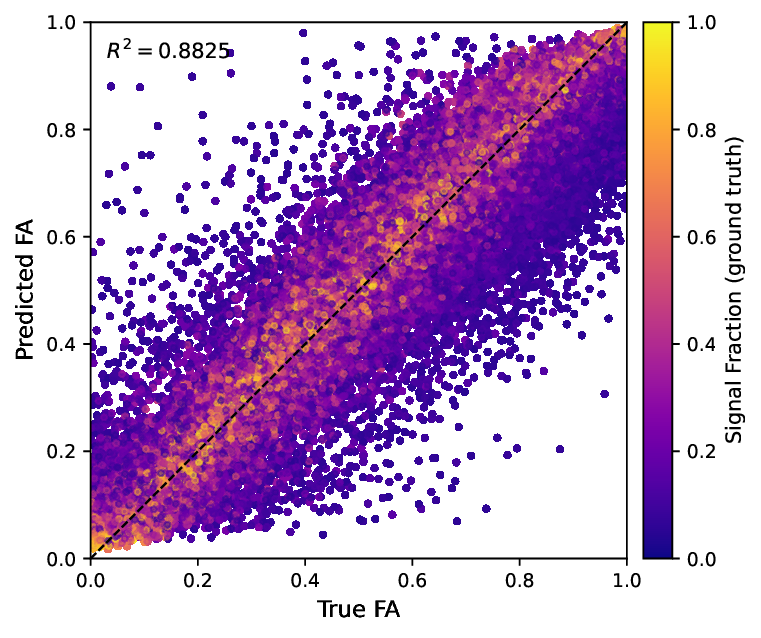}
        \caption{Predicted vs. true FA}
        \label{sfig:scatter_fa}
    \end{subfigure}
    \par\medskip
    \begin{subfigure}[t]{0.37\textwidth}
        \centering
        \includegraphics[width=\linewidth]{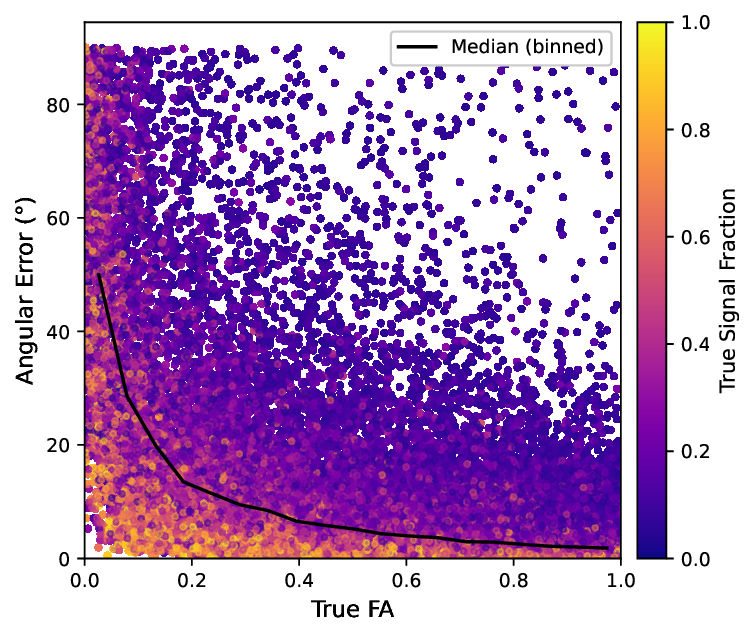}
        \caption{Angular error vs. FA}
        \label{sfig:scatter_angular}
    \end{subfigure}
    \hspace{0.015\textwidth}
    \begin{subfigure}[t]{0.37\textwidth}
        \centering
        \includegraphics[width=\linewidth]{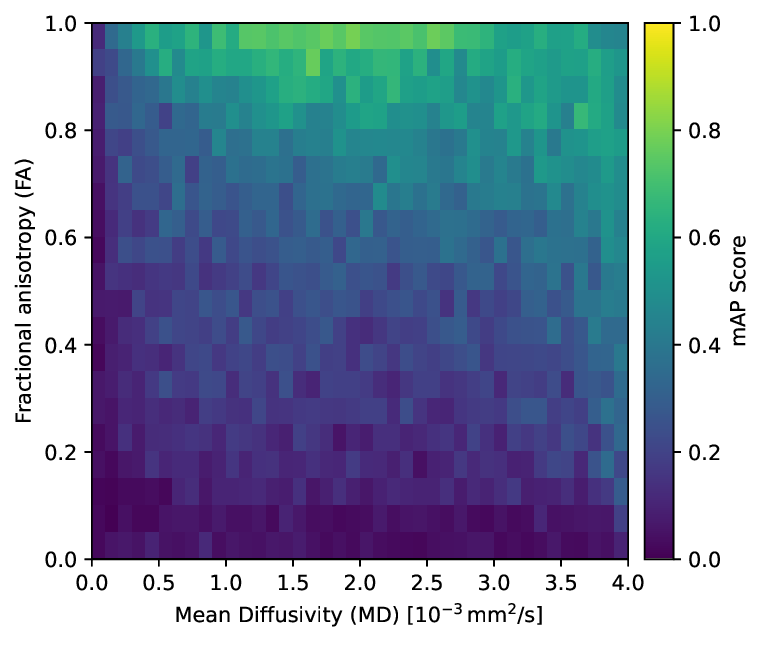}
        \caption{mAP@10 map}
        \label{sfig:map_heatmap}
    \end{subfigure}
    \caption{Quantitative evaluation of our test data. (a,\,b)~True vs. predicted MD and FA for all Hungarian-matched compartments with existence score $\geq0.35$, colored by signal fraction. (c)~Angular error of the predicted principal direction vs. true FA, colored by signal fraction. Solid line shows the binned median angular error. (d)~Mean average precision mAP@10 across the MD/FA parameter space.}
    \label{fig:eval}
\end{figure}

Fig. \ref{fig:eval} shows different aspects of our quantitative evaluation. Scatter plots show \ref{sfig:scatter_md}: predicted vs. true MD, \ref{sfig:scatter_fa}: predicted vs. true FA, and \ref{sfig:scatter_angular} angular error vs. true FA. Here, compartmental signal fractions are color coded. The coefficient of determination is $R^2=0.9469$ for MD, and $R^2=0.8825$ for FA. Further we observe a median angular error of $\SI{4.24}{\degree}$. All scatter plots clearly show that compartments with higher signal fraction are reconstructed more reliably than compartments with low weight. Additionally, Fig. \ref{sfig:scatter_angular} shows that the prediction of main directions is more accurate for higher FA with errors rising sharply for $FA\leq0.2$. Fig. \ref{sfig:map_heatmap} shows the mAP@10 heatmap in the MD-FA space, clearly indicating that compartments with high FA are reconstructed more precisely than low FA compartments. This trend is similar albeit less expressed for MD.

\begin{table}
    \centering
    \caption{mAP@10 scores computed for different groups of compartments are shown. Compartments are grouped by weight and classified as either small, medium or large based on the thresholds defined in the table.}\label{tab:map}
    \small 
    \setlength{\tabcolsep}{5pt} 
    \renewcommand{\arraystretch}{1.3}
    \begin{tabular}{rcr}
        \hline
        \textbf{Compartment size} & \textbf{mAP@10} & \textbf{count} \\ 
        \hline
        small ($0.05 \leq f < 0.35$) & 0.142 & 61249\\
        medium ($0.35 \leq f < 0.65$) & 0.562 & 20065\\
        large ($0.65 \leq f \leq 0.95$) & 0.766 & 6186\\
        \hline
    \end{tabular}
\end{table}

\section{Discussion}
We propose a new way to approach 5D diffusion correlation imaging that integrates information about diffusivity (MD), anisotropy (FA), and fiber directions. To reconstruct multiple diffusion tensors, i.e. compartments, without solving a highly ill-conditioned ILT to recover a full 5D spectrum, we adapt object detection principles. Our network is based on DETR \cite{carion2020}, but both architecture and loss functions are adapted to work well with our multi-tensor diffusion data. Additionally, mean Average Precision is introduced as an intuitive metric to benchmark model performance and compare our results with future developments.

The model performance highly depends on signal fraction, as compartments with little weight have little influence on overall signal. This automatically makes the reconstruction of samples with many compartments challenging. Higher MD and especially higher FA result in more reliable reconstructions, likely because high MD/FA cause more dynamic signal curves. Generally, both MD and FA are estimated quite accurately over the full range of values. Main direction is estimated accurately for moderate to high FA, with correct estimation being unrealistic and unnecessary at low FA. Generally, residual errors in estimation of diffusion parameters given noisy measurement are to be expected \cite{basser_noise,jones_noise}.

Compared to nonparametric correlation imaging studies that recover similarly rich outputs including fiber orientation \cite{dealmeidamartins2021computing,martin2021nonparametric,reymbaut2021toward,Rosenberg2022}, our approach does not require specialized sequences with multiple b-tensor shapes, relying instead on standard multi-shell/HARDI acquisitions. Unlike fiber orientation methods \cite{behrens2007probabilistic,tournier2007robust,JEURISSEN2014411}, we simultaneously recover compartmental DTI metrics without a separate microstructure estimation step. And unlike parametric models \cite{zhang2012noddi,scherrer2016characterizing}, we impose no fixed assumptions on compartment type or count. Together, this positions our work in a unique niche: joint, assumption-free microstructure and fiber orientation quantification from a standard acquisition.

There are several limitations to this study. While the use of synthetic data allows us to objectively assess model performance, generalization to in vivo data remains to be tested, as there are several effects in vivo that are currently not modeled by our pipeline, such as Rician noise, non-Gaussian diffusion, flow, or Magnetization Transfer. Furthermore, the identifiability of multi-compartment diffusion tensors from multi-shell acquisitions with linear b-tensors is limited: it has been shown that infinitely many compartment configurations produce identical single-shell signals, and that an extension to multiple b-values only reduces but does not eliminate this degeneracy \cite{taquet2013,jelescu2016}. Our network implicitly addresses this through learned priors embedded in the synthetic data \cite{DEALMEIDAMARTINS2021118601}, analogously to the population-informed prior proposed by Taquet et al. \cite{taquet2013} — but the extent to which this resolves degeneracy under realistic in vivo noise conditions remains to be validated. Introducing stronger or more targeted priors, for instance by adapting the training data distribution to reflect realistic tissue parameter ranges \cite{gyori2022} or by incorporating spatial regularization, is a promising direction to further reduce degeneracy. To that end, we plan to leverage similarity in voxels of the same structure or connected fibers. Other post-processing could include a refinement of the predicted compartments, e.g. refitting of weights to the actual signal curve. Compared to our initial proof-of-concept, we already reduced the number of inputs by $\SI{50}{\percent}$ and still increased model performance substantially \cite{endt2026}. However, further reduction of required contrasts is needed to allow clinical application.

\section{Conclusion}
We reframe compartmental microstructure quantification as an object detection-like problem, enabling joint recovery of MD, FA, main fiber direction, and signal fraction for a variable number of compartments per voxel from standard multi-shell diffusion MRI. By adapting the DETR architecture and training with Hungarian matching, we avoid both the ill-posed inverse Laplace transform underlying conventional spectrum reconstruction and the fixed-compartment assumptions of parametric tissue models. Results on a large synthetic dataset demonstrate reliable recovery of dominant compartments across voxel configurations with up to five sub-compartments, while the accuracy for small compartments decreases, as their signal contribution is limited. Our approach fills a methodological gap left by existing approaches: neither fiber orientation methods nor parametric microstructure models jointly recover the full set of compartmental parameters from linear diffusion encoding alone. In vivo validation and the incorporation of spatial context remain the critical next steps toward clinical translation.
\begin{credits}

\subsubsection{Code availability}
All code to reproduce our work with defined seeds is publicly available at \github.


\subsubsection{\discintname}
The authors have no competing interests to declare that are relevant to the content of this article.
\end{credits}
%
\bibliographystyle{splncs04}
\bibliography{biblio}

@inproceedings{schlund2025,
  author = {Schlund, Johannes R. and Endt, Sebastian and Menzel, Marion I.},
  title = {A deep learning approach for direction-aware diffusion-diffusion correlation imaging},
  booktitle={European Society for Magnetic Resonance in Medicine and Biology (ESMRMB) 2025 41st Annual Scientific Meeting},
  organization = {Magnetic Resonance Materials in Physics, Biology and Medicine},
  volume={38},
  pages={908--909},
  year = 2025,
  doi = {10.1007/s10334-025-01300-z}
}

@inproceedings{endt2026,
  author = {Endt, Sebastian and Schlund, Johannes R. and Wirth, Marcus and Menzel, Marion I.},
  title = {{Detection Transformer for Direction-Aware Diffusion-Diffusion Correlation Imaging}},
  booktitle={European Society for Magnetic Resonance in Medicine and Biology (ESMRMB) 2026 42nd Annual Scientific Meeting},
  organization = {Magnetic Resonance Materials in Physics, Biology and Medicine},
  volume={39},
  month={Oct},
  year = 2026,
  note = {(Accepted)}
}

@misc{hansendata,
  author = {Hansen, Brian and Jespersen, Sune Nørhøj},
  year = 2017,
  title = {{Data for evaluation of fast kurtosis imaging, b-value optimization and exploration of diffusion MRI contrast [Dataset]}},
  howpublished = {Dryad},
  doi = {10.5061/dryad.9bc43}
}

@article{hansen2016,
  title={Data for evaluation of fast kurtosis strategies, b-value optimization and exploration of diffusion {MRI} contrast},
  author={Hansen, Brian and Jespersen, Sune N{\o}rh{\o}j},
  journal={Scientific data},
  volume={3},
  number={1},
  pages={1--5},
  year={2016},
  publisher={Nature Publishing Group},
  doi={10.1038/sdata.2016.72}
}

@InProceedings{carion2020,
  author="Carion, Nicolas and Massa, Francisco and Synnaeve, Gabriel and Usunier, Nicolas and Kirillov, Alexander and Zagoruyko, Sergey",
  title="End-to-End Object Detection with Transformers",
  booktitle="Computer Vision -- ECCV 2020",
  year="2020",
  publisher="Springer International Publishing",
  address="Cham",
  pages="213--229",
  isbn="978-3-030-58452-8",
  doi={10.1007/978-3-030-58452-8_13}
}

@inproceedings{endt2021unmixing,
  title={{Unmixing tissue compartments via deep learning T1-T2-relaxation correlation imaging}},
  author={Endt, Sebastian and Pirkl, Carolin M and Verdun, Claudio M and Menze, Bjoern H and Menzel, Marion I},
  booktitle={17th International Symposium on Medical Information Processing and Analysis},
  volume={12088},
  pages={218--227},
  year={2021},
  organization={SPIE},
  doi={10.1117/12.2604737}
}

@article{kim2020,
  title={Multidimensional correlation spectroscopic imaging of exponential decays: from theoretical principles to in vivo human applications},
  author={Kim, Daeun and Wisnowski, Jessica L. and Nguyen, Christopher T. and Haldar, Justin P.},
  journal={NMR in Biomedicine},
  volume={33},
  number={12},
  pages={e4244},
  year={2020},
  doi={10.1002/nbm.4244}
}

@article{yu2021,
  title={Model-informed machine learning for multi-component {{T2}} relaxometry},
  author={Yu, T. and Canales-Rodr{\'\i}guez, E.J. and Pizzolato, M. and Piredda, G.F. and Hilbert, T. and Fischi-Gomez, E. and Weigel, M. and Barakovic, M. and Cuadra, M.B. and Granziera, C. and Kober, T. and Thiran J.-P.},
  journal={Medical Image Analysis},
  volume={69},
  pages={101940},
  year={2021},
  publisher={Elsevier},
  doi={10.1016/j.media.2020.101940}
}

@article{Avram.2021,
 author = {Avram, Alexandru V. and Sarlls, Joelle E. and Basser, Peter J.},
 year = {2021},
 title = {{Whole-Brain Imaging of Subvoxel T1-Diffusion Correlation Spectra in Human Subjects}},
 pages = {671465},
 volume = {15},
 journal = {{Frontiers in neuroscience}},
 doi = {10.3389/fnins.2021.671465}
}

@article{martin2021nonparametric,
    title={{Nonparametric D-R1-R2 distribution MRI of the living human brain}},
  author={Martin, Jan and Reymbaut, Alexis and Schmidt, Manuel and Doerfler, Arnd and Uder, Michael and Laun, Frederik Bernd and Topgaard, Daniel},
  journal={NeuroImage},
  volume={245},
  pages={118753},
  year={2021},
  doi={10.1016/j.neuroimage.2021.118753}
}

@article{reymbaut2021toward,
    title={Toward nonparametric diffusion-characterization of crossing fibers in the human brain},
  author={Reymbaut, Alexis and Critchley, Jeffrey and Durighel, Giuliana and Sprenger, Tim and Sughrue, Michael and Bryskhe, Karin and Topgaard, Daniel},
  journal={Magnetic Resonance in Medicine},
  volume={85},
  number={5},
  pages={2815--2827},
  year={2021},
  doi={10.1002/mrm.28604}
}

@article{dealmeidamartins2021computing,
    title={Computing and visualising intra-voxel orientation-specific relaxation--diffusion features in the human brain},
  author={de Almeida Martins, Jo{\~a}o P and Tax, Chantal M. W. and Reymbaut, Alexis and Szczepankiewicz, Filip and Chamberland, Maxime and Jones, Derek K. and Topgaard, Daniel},
  journal={Human brain mapping},
  volume={42},
  number={2},
  pages={310--328},
  year={2021},
  doi={10.1002/hbm.25224},
  publisher={Wiley Online Library}
}

@article{Rosenberg2022,
  author = {Rosenberg, Jens T. and Grant, Samuel C. and Topgaard, Daniel},
 year = {2022},
 title = {{Nonparametric 5D D-R2 distribution imaging with single-shot EPI at 21.1 T: Initial results for in vivo rat brain}},
 pages = {107256},
 volume = {341},
 journal = {Journal of magnetic resonance (San Diego, Calif. : 1997)},
 doi = {10.1016/j.jmr.2022.107256}
}

@article{Luo.2023,
 author = {Luo, P. and Hu, W. and Xu, R. and Wang, Y. and Li, X. and Jiang, L. and Chang, S. and Wu, D. and Li, G. and Dai, Y.},
 year = {2023},
 title = {{Enabling early detection of knee osteoarthritis using diffusion-relaxation correlation spectrum imaging}},
 pages = {e681--e687},
 volume = {78},
 number = {9},
 journal = {{Clinical radiology}},
 doi = {10.1016/j.crad.2023.05.013}
}

@article{MacKay.1994,
 author = {MacKay, A. and Whittall, K. and Adler, J. and Li, D. and Paty, D. and Graeb, D.},
 year = {1994},
 title = {{In vivo visualization of myelin water in brain by magnetic resonance}},
 pages = {673--677},
 volume = {31},
 number = {6},
 issn = {0740-3194},
 journal = {{Magnetic Resonance in Medicine}},
 doi = {10.1002/mrm.1910310614}
}

@article{Naranjo.2021,
 author = {Naranjo, Isaac Daimiel and Reymbaut, Alexis and Brynolfsson, Patrik and {Lo Gullo}, Roberto and Bryskhe, Karin and Topgaard, Daniel and Giri, Dilip D. and Reiner, Jeffrey S. and Thakur, Sunitha B. and Pinker-Domenig, Katja},
 year = {2021},
 title = {{Multidimensional Diffusion Magnetic Resonance Imaging for Characterization of Tissue Microstructure in Breast Cancer Patients: A Prospective Pilot Study}},
 volume = {13},
 number = {7},
 issn = {2072-6694},
 journal = {{Cancers}},
 doi = {10.3390/cancers13071606}
}

@article{Ning.2025,
 author = {Ning, Lipeng},
 year = {2025},
 title = {{Maximum-entropy and subspace methods for high-resolution relaxation-diffusion distribution estimation}},
 volume = {3},
 journal = {{Imaging neuroscience (Cambridge, Mass.)}},
 doi = {10.1162/IMAG.a.113}
}

@article{Xu.2024,
 author = {Xu, Junqi and Sheng, Yaru and Li, Hao and Yang, Zidong and Ren, Yan and Wang, He},
 year = {2024},
 title = {{A data-driven intravoxel mean diffusivities distribution approach for molecular classifications and MIB-1 prediction of gliomas}},
 pages = {7332--7344},
 volume = {51},
 number = {10},
 journal = {{Medical physics}},
 doi = {10.1002/mp.17280}
}

@article{Liu.2024,
author = {Liu, Qiang and Gagoski, Borjan and Shaik, Imam Ahmed and Westin, Carl-Fredrik and Wilde, Elisabeth A. and Schneider, Walter and Bilgic, Berkin and Grissom, William A. and Nielsen, Jon-Fredrik and Zaitsev, Maxim and Rathi, Yogesh and Ning, Lipeng},
 year = {2024},
 title = {{Time-division multiplexing (TDM) sequence removes bias in T2 estimation and relaxation-diffusion measurements}},
 pages = {2506--2519},
 volume = {92},
 number = {6},
 journal = {{Magnetic Resonance in Medicine}},
 doi = {10.1002/mrm.30246}
}

@article{Slator.2019,
 author = {Slator, Paddy J. and Hutter, Jana and Palombo, Marco and Jackson, Laurence H. and Ho, Alison and Panagiotaki, Eleftheria and Chappell, Lucy C. and Rutherford, Mary A. and Hajnal, Joseph V. and Alexander, Daniel C.},
 year = {2019},
 title = {{Combined diffusion-relaxometry MRI to identify dysfunction in the human placenta}},
 pages = {95--106},
 volume = {82},
 number = {1},
 journal = {{Magnetic Resonance in Medicine}},
 doi = {10.1002/mrm.27733}
}

@article{Avram.2019,
 author = {Avram, Alexandru V. and Sarlls, Joelle E. and Basser, Peter J.},
 year = {2019},
 title = {{Measuring non-parametric distributions of intravoxel mean diffusivities using a clinical MRI scanner}},
 pages = {255--262},
 volume = {185},
 journal = {{NeuroImage}},
 doi = {10.1016/j.neuroimage.2018.10.030}
}

@article{Hu.2024,
 author = {Hu, Wentao and Dai, Yongming and Liu, Fang and Yang, Tianshu and Wang, Yao and Shen, Yiwei and Zhou, Wenyan and Wu, Dongmei and Gu, Leyi and Zhang, Minfang and Zhou, Yan},
 year = {2024},
 title = {{Assessing renal interstitial fibrosis using compartmental, non-compartmental, and model-free diffusion MRI approaches}},
 pages = {156},
 volume = {15},
 number = {1},
 journal = {{Insights into imaging}},
 doi = {10.1186/s13244-024-01736-2}
}

@article{Endt.2023,
 author = {Endt, Sebastian and Engel, Maria and Naldi, Emanuele and Assereto, Rodolfo and Molendowska, Malwina and Mueller, Lars and {Mayrink Verdun}, Claudio and Pirkl, Carolin M. and Palombo, Marco and Jones, Derek K. and Menzel, Marion I.},
 year = {2023},
 title = {{In Vivo Myelin Water Quantification Using Diffusion-Relaxation Correlation MRI: A Comparison of 1D and 2D Methods}},
 pages = {1571--1588},
 volume = {54},
 number = {11-12},
 journal = {{Applied magnetic resonance}},
 doi = {10.1007/s00723-023-01584-1}
}

@article{Zong.2024,
 author = {Zong, Fangrong and Wang, Lixian and Liu, Huabing and Xue, Bing and Bai, Ruiliang and Liu, Yong},
 year = {2024},
 title = {{A genetic optimisation and iterative reconstruction framework for sparse multi-dimensional diffusion-relaxation correlation MRI}},
 pages = {108508},
 volume = {175},
 journal = {{Computers in biology and medicine}},
 doi = {10.1016/j.compbiomed.2024.108508}
}

@article{Nagtegaal.2023,
 author = {Nagtegaal, Martijn and Hartsema, Emiel and Koolstra, Kirsten and Vos, Frans},
 year = {2023},
 title = {{Multicomponent MR fingerprinting reconstruction using joint-sparsity and low-rank constraints}},
 pages = {286--298},
 volume = {89},
 number = {1},
 issn = {0740-3194},
 journal = {{Magnetic Resonance in Medicine}},
 doi = {10.1002/mrm.29442}
}

@article{jeurissen2013investigating,
  title={Investigating the prevalence of complex fiber configurations in white matter tissue with diffusion magnetic resonance imaging},
  author={Jeurissen, Ben and Leemans, Alexander and Tournier, Jacques-Donald and Jones, Derek K and Sijbers, Jan},
  journal={Human brain mapping},
  volume={34},
  number={11},
  pages={2747--2766},
  year={2013},
  publisher={Wiley Online Library},
  doi = {10.1002/hbm.22099}
}

@article{zhang2012noddi,
  title={{NODDI: practical in vivo neurite orientation dispersion and density imaging of the human brain}},
  author={Zhang, Hui and Schneider, Torben and Wheeler-Kingshott, Claudia A and Alexander, Daniel C},
  journal={NeuroImage},
  volume={61},
  number={4},
  pages={1000--1016},
  year={2012},
  publisher={Elsevier},
  doi={10.1016/j.neuroimage.2012.03.072}
}

@article{scherrer2016characterizing,
  title={{Characterizing brain tissue by assessment of the distribution of anisotropic microstructural environments in diffusion-compartment imaging (DIAMOND)}},
  author={Scherrer, Benoit and Schwartzman, Armin and Taquet, Maxime and Sahin, Mustafa and Prabhu, Sanjay P and Warfield, Simon K},
  journal={Magnetic resonance in medicine},
  volume={76},
  number={3},
  pages={963--977},
  year={2016},
  publisher={Wiley Online Library},
  doi={10.1002/mrm.25912}
}

@article{tournier2007robust,
  title={{Robust determination of the fibre orientation distribution in diffusion MRI: non-negativity constrained super-resolved spherical deconvolution}},
  author={Tournier, J-Donald and Calamante, Fernando and Connelly, Alan},
  journal={NeuroImage},
  volume={35},
  number={4},
  pages={1459--1472},
  year={2007},
  publisher={Elsevier},
  doi={10.1016/j.neuroimage.2007.02.016}
}

@article{JEURISSEN2014411,
title = {{Multi-tissue constrained spherical deconvolution for improved analysis of multi-shell diffusion MRI data}},
journal = {NeuroImage},
volume = {103},
pages = {411--426},
year = {2014},
issn = {1053-8119},
doi = {10.1016/j.neuroimage.2014.07.061},
author = {Ben Jeurissen and Jacques-Donald Tournier and Thijs Dhollander and Alan Connelly and Jan Sijbers}
}

@article{behrens2007probabilistic,
  title={Probabilistic diffusion tractography with multiple fibre orientations: What can we gain?},
  author={Behrens, Timothy EJ and Berg, H Johansen and Jbabdi, Saad and Rushworth, Matthew FS and Woolrich, Mark W},
  journal={NeuroImage},
  volume={34},
  number={1},
  pages={144--155},
  year={2007},
  publisher={Elsevier},
  doi={10.1016/j.neuroimage.2006.09.018}
}

@article{pascalvoc,
author = {Everingham, Mark and Van Gool, Luc and Williams, Christopher and Winn, John and Zisserman, Andrew},
year = {2010},
month = {06},
pages = {303--338},
title = {The Pascal Visual Object Classes (VOC) challenge},
volume = {88},
journal = {International Journal of Computer Vision},
doi = {10.1007/s11263-009-0275-4}
}

@inproceedings{mscoco,
	address = {Cham},
	title = {Microsoft {COCO}: {Common} {Objects} in {Context}},
	isbn = {978-3-319-10602-1},
	shorttitle = {Microsoft {COCO}},
	doi = {10.1007/978-3-319-10602-1_48},
	booktitle = {Computer {Vision} – {ECCV} 2014},
	publisher = {Springer International Publishing},
	author = {Lin, Tsung-Yi and Maire, Michael and Belongie, Serge and Hays, James and Perona, Pietro and Ramanan, Deva and Dollár, Piotr and Zitnick, C. Lawrence},
	editor = {Fleet, David and Pajdla, Tomas and Schiele, Bernt and Tuytelaars, Tinne},
	year = {2014},
	pages = {740--755}
}

@inproceedings{Hungarian,
  author={Stewart, Russell and Andriluka, Mykhaylo and Ng, Andrew Y.},
  booktitle={2016 IEEE Conference on Computer Vision and Pattern Recognition (CVPR)}, 
  title={End-to-End People Detection in Crowded Scenes}, 
  year={2016},
  pages={2325--2333},
  doi={10.1109/CVPR.2016.255}
}

@article{focal_loss,
  author  = {Lin, Tsung-Yi and Goyal, Priya and Girshick, Ross and He, Kaiming and Dollár, Piotr},
  journal = {IEEE Transactions on Pattern Analysis and Machine Intelligence},
  title   = {Focal Loss for Dense Object Detection},
  year    = {2020},
  volume  = {42},
  number  = {2},
  pages   = {318--327},
  doi     = {10.1109/TPAMI.2018.2858826}
}

@article{basser_noise,
    author = {Basser, Peter J. and Pajevic, Sinisa},
    title = {{Statistical artifacts in diffusion tensor MRI (DT-MRI) caused by background noise}},
    journal = {Magnetic Resonance in Medicine},
    volume = {44},
    number = {1},
    pages = {41--50},
    doi = {10.1002/1522-2594(200007)44:1<41::AID-MRM8>3.0.CO;2-O},
    year = {2000}
}

@article{jones_noise,
author = {Jones, Derek K. and Basser, Peter J.},
title = {{“Squashing peanuts and smashing pumpkins”: How noise distorts diffusion-weighted MR data}},
journal = {Magnetic Resonance in Medicine},
volume = {52},
number = {5},
pages = {979--993},
doi = {10.1002/mrm.20283},
year = {2004}
}

@article{BOGUSZ2025110998,
title = {{Multi-compartment diffusion–relaxation MR signal representation in the spherical 3D-SHORE basis}},
journal = {Computers in Biology and Medicine},
volume = {197},
pages = {110998},
year = {2025},
doi = {10.1016/j.compbiomed.2025.110998},
author = {Fabian Bogusz and Tomasz Pieciak}
}

@InProceedings{taquet2013,
author={Taquet, Maxime and Scherrer, Beno{\^i}t and Boumal, Nicolas and Macq, Beno{\^i}t and Warfield, Simon K.},
title={{Estimation of a Multi-fascicle Model from Single B-Value Data with a Population-Informed Prior}},
booktitle={Medical Image Computing and Computer-Assisted Intervention -- MICCAI 2013},
year={2013},
pages={695--702},
doi={10.1007/978-3-642-40811-3_87}
}

@article{jelescu2016,
author = {Jelescu, Ileana O. and Veraart, Jelle and Fieremans, Els and Novikov, Dmitry S.},
title = {Degeneracy in model parameter estimation for multi-compartmental diffusion in neuronal tissue},
journal = {{NMR in Biomedicine}},
volume = {29},
number = {1},
pages = {33-47},
doi = {10.1002/nbm.3450},
year = {2016}
}

@article{DEALMEIDAMARTINS2021118601,
title = {Neural networks for parameter estimation in microstructural MRI: Application to a diffusion-relaxation model of white matter},
journal = {NeuroImage},
volume = {244},
pages = {118601},
year = {2021},
issn = {1053-8119},
doi = {10.1016/j.neuroimage.2021.118601},
author = {João P. {de Almeida Martins} and Markus Nilsson and Björn Lampinen and Marco Palombo and Peter T. While and Carl-Fredrik Westin and Filip Szczepankiewicz}
}

@article{REYMBAUT2021101988,
title = {Magic {DIAMOND}: Multi-fascicle diffusion compartment imaging with tensor distribution modeling and tensor-valued diffusion encoding},
journal = {Medical Image Analysis},
volume = {70},
pages = {101988},
year = {2021},
doi = {10.1016/j.media.2021.101988},
author = {Alexis Reymbaut and Alex Valcourt Caron and Guillaume Gilbert and Filip Szczepankiewicz and Markus Nilsson and Simon K. Warfield and Maxime Descoteaux and Benoit Scherrer}
}

@article{reisert2017,
title = {{Disentangling micro from mesostructure by diffusion MRI: A Bayesian approach}},
journal = {NeuroImage},
volume = {147},
pages = {964-975},
year = {2017},
doi = {10.1016/j.neuroimage.2016.09.058},
author = {Marco Reisert and Elias Kellner and Bibek Dhital and Jürgen Hennig and Valerij G. Kiselev}
}

@ARTICLE{dessain2024,
AUTHOR={Dessain, Quentin  and Fuchs, Clément  and Macq, Benoît  and Rensonnet, Gaëtan },
TITLE={Fast multi-compartment Microstructure Fingerprinting in brain white matter},
JOURNAL={Frontiers in Neuroscience},
VOLUME={18},
YEAR={2024},
DOI={10.3389/fnins.2024.1400499}
}

@article{consagra2025,
title = {A deep learning approach to multi-fiber parameter estimation and uncertainty quantification in diffusion MRI},
journal = {Medical Image Analysis},
volume = {102},
pages = {103537},
year = {2025},
doi = {10.1016/j.media.2025.103537},
author = {William Consagra and Lipeng Ning and Yogesh Rathi}
}

@ARTICLE{golkov2016,
  author={Golkov, Vladimir and Dosovitskiy, Alexey and Sperl, Jonathan I. and Menzel, Marion I. and Czisch, Michael and Sämann, Philipp and Brox, Thomas and Cremers, Daniel},
  journal={IEEE Transactions on Medical Imaging}, 
  title={{q-Space Deep Learning: Twelve-Fold Shorter and Model-Free Diffusion MRI Scans}}, 
  year={2016},
  volume={35},
  number={5},
  pages={1344-1351},
  doi={10.1109/TMI.2016.2551324}
}

@article{gyori2022,
author = {Gyori, Noemi G. and Palombo, Marco and Clark, Christopher A. and Zhang, Hui and Alexander, Daniel C.},
title = {Training data distribution significantly impacts the estimation of tissue microstructure with machine learning},
journal = {Magnetic Resonance in Medicine},
volume = {87},
number = {2},
pages = {932-947},
doi = {10.1002/mrm.29014},
year = {2022}
}
\end{document}